\documentclass{article} 
\usepackage{iclr2027_conference,times}

\usepackage{amsmath,amsfonts,bm}

\def\eqref#1{equation~\ref{#1}}

\def\1{\bm{1}}

\DeclareMathAlphabet{\mathsfit}{\encodingdefault}{\sfdefault}{m}{sl}
\SetMathAlphabet{\mathsfit}{bold}{\encodingdefault}{\sfdefault}{bx}{n}

\usepackage{amsthm}
\usepackage{graphicx}
\usepackage{booktabs}
\usepackage{wrapfig}
\usepackage{placeins}
\usepackage{xcolor}
\usepackage{hyperref}
\usepackage{url}
\usepackage{algorithm,algpseudocode}

\title{RaMMOBO: Random Max-Margin \\ Multi-Objective Bayesian Optimization}

\author{Thomas Wedenig$^{1,2}$, Paul Plantosar$^{1,2}$, Robert Peharz$^{1,2}$ \\
$^{1}$Institute of Machine Learning and Neural Computation, TU Graz, Austria \\
$^{2}$Graz Center for Machine Learning (GraML), Austria \\
\texttt{\{name.surname\}@tugraz.at}
}

\newcommand{\chg}[1]{#1}

\iclrfinalcopy 
\begin{document}

\maketitle

\begin{abstract}

Multi-objective Bayesian optimization (MOBO) maximizes $M$ expensive black-box objectives with the goal of approximating the \emph{Pareto set}, whose objective vectors cannot be improved in one component without sacrificing another.
A common quality measure is \emph{hypervolume}, the volume in objective space dominated by these vectors relative to a reference point.
\emph{Expected hypervolume improvement} (EHVI) directly targets this metric and is one-step Bayes-optimal, but its standard computation scales exponentially with $M$, making it prohibitively expensive for many-objective problems.
We introduce \emph{random max-margin multi-objective Bayesian optimization} ({RaMMOBO}), whose acquisition function lower-bounds EHVI while requiring only linear cost in $M$ to evaluate.
RaMMOBO measures the volume of a max-margin box contained in a candidate's hypervolume-improvement region and averages over random reweightings of the objective space, yielding a utility that is strictly Pareto-monotone on hypervolume-improving outcomes.
It naturally handles noisy observations and different probabilistic surrogates and supports efficient gradient-based optimization.
Across synthetic and real-world benchmarks, RaMMOBO matches or outperforms state-of-the-art methods, with largest gains at higher $M$, with acquisition optimization up to two orders of magnitude faster than $q$LogNEHVI.
Our implementation is available at \url{https://anonymous.4open.science/r/rammobo}.

\end{abstract}

\section{Introduction}

\label{sec:intro}

Optimizing expensive scalar-valued black-box functions is central to scientific discovery, engineering design, and machine learning.
Bayesian optimization (BO) addresses this problem by fitting a probabilistic surrogate to the observed data and using an acquisition function to decide where to query the function next \citep{shahriari2016taking}.
Multi-objective Bayesian optimization (MOBO) extends BO to problems with $M$ competing objectives. In this setting, there is generally no single optimal solution: improving one objective may require worsening another. The goal is therefore to efficiently identify a set of Pareto-optimal solutions that represents different trade-offs between the objectives. Such problems arise, for example, in drug discovery and engineering design, and many practical applications may involve substantially more than three objectives \citep{fleming2005many,li2015many,fromer2023computer,tanabe2020re}.

As $M$ grows, however, selecting the next evaluation can itself become a computational bottleneck.
Effective MOBO therefore requires acquisition functions that guide the search effectively while remaining computationally efficient to evaluate and optimize. 
Hypervolume-based acquisitions such as (noisy) expected hypervolume improvement ((N)EHVI) directly quantify improvement in dominated volume and are known to be one-step Bayes-optimal for hypervolume maximization \citep{daulton2021parallel}.
However, standard implementations rely on geometric decompositions of the objective space
\citep{daulton2021parallel,ament2023unexpected}, 
whose computational cost grows exponentially in $M$ \citep{yang2019efficient}.
Hence, for problems with many objectives, hypervolume-based acquisition functions can quickly become computationally prohibitive to evaluate.
Many alternative acquisition functions replace hypervolume computation with other costly subroutines: information-theoretic and posterior-sampling methods often require auxiliary multi-objective optimization
\citep{hernandezlobato2016predictive,belakaria2019maxvalue,tu2022joint,renganathan2025qpots},
while BOtied adds multivariate CDF fitting and gradient-free acquisition optimization
\citep{park2024botied}.
Scalarization is computationally cheaper, but its Pareto-front coverage depends strongly on the choice of scalarization and weight distribution
\citep{knowles2006parego,zhang2020random}.
These trade-offs motivate an acquisition function that retains a direct
connection to multi-objective improvement while avoiding box decompositions
and costly auxiliary optimization.

\begin{wrapfigure}{R}{0.4\textwidth}
    \includegraphics[width=\linewidth]{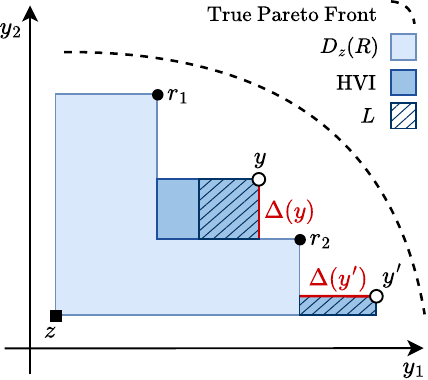}
    \caption{RaMMOBO's utility $L$ lower-bounds hypervolume improvement (HVI); see text for details.}
    \label{fig:rammobo-vs-hvi}
\end{wrapfigure}

We introduce \textbf{RaMMOBO}, \emph{random max-margin multi-objective
Bayesian optimization}, a simple alternative to hypervolume-based acquisition
functions that avoids box decompositions altogether. Consider Figure~\ref{fig:rammobo-vs-hvi}, which shows the objective space of \chg{an} $M=2$\chg{-}dimensional maximization problem with current Pareto-front approximation $R = \{r_1, r_2\}$, which dominates the space $D_z(R)$ below it, up to a reference point $z$.
Rather than explicitly computing \emph{hypervolume improvement} (HVI) for a candidate objective vector $y
$, RaMMOBO intuitively measures the shortest axis-aligned distance to the region dominated by $R$, called the margin $\Delta(y)$. RaMMOBO's utility $L(y)$ then measures the volume of a hypercube ending at $y$ with side length $\Delta(y)$---while clipping the cube to a box for objective vectors like $y'$, where the cube would include vectors that are worse than the reference $z$ in some dimension.
With this construction, $L$ gives a direct lower bound on HVI without costly partitions of the objective space. 

Furthermore, akin to random scalarization \citep{knowles2006parego}, RaMMOBO randomizes the relative axis scalings of objective space and averages the resulting volumes, which yields a utility that is strictly Pareto-monotone on objective vectors that improve hypervolume.
Crucially, for $N = |R|$ nondominated points, evaluating RaMMOBO's utility requires only
$\mathcal{O}(NM)$ operations, while hypervolume-based methods scale exponentially in $M$.
RaMMOBO naturally incorporates noisy observations, can be used with any surrogate model that provides differentiable posterior samples, and its acquisition function can be efficiently optimized via gradient-based methods.

We evaluate RaMMOBO on synthetic and real-world benchmarks spanning $M=2$ to $14$ objectives, including noisy and batch settings.
RaMMOBO is competitive at lower objective counts and becomes increasingly advantageous as $M$ grows, with acquisition optimization roughly two orders of magnitude faster than common methods based on hypervolume.

In summary, our contributions are threefold:
\begin{itemize}
    \item We introduce RaMMOBO, whose evaluation cost scales linearly in the number of objectives $M$, avoiding the expensive box decompositions required by hypervolume-based methods.
    RaMMOBO naturally supports noisy observations and different probabilistic surrogates and admits efficient gradient-based acquisition optimization.

    \item We show that RaMMOBO retains key properties of hypervolume improvement: its utility is strictly Pareto-monotone on hypervolume-improving outcomes, and its acquisition lower-bounds noisy expected hypervolume improvement.

    \item We demonstrate strong empirical performance on synthetic and real-world benchmarks in sequential and batch settings.
    RaMMOBO is competitive at $M=8$ and achieves state-of-the-art performance for $M\geq10$, while reducing acquisition evaluation time by more than two orders of magnitude compared with hypervolume-based methods.
\end{itemize}

\section{Background}

\subsection{Bayesian optimization}

Bayesian optimization (BO) optimizes an expensive scalar-valued black-box function $f \colon \mathcal X\to\mathbb R$ under a limited evaluation budget by maintaining a probabilistic surrogate, often a Gaussian process (GP) \citep{rasmussen2006gaussian}.
Given noiseless observations $\mathcal D=\{(x_i,y_i)\}_{i=1}^n$ with $y_i=f(x_i)$, the surrogate represents uncertainty about $f(x)$ through a random objective value $Y(x)$ with posterior predictive distribution $p(Y(x)\mid\mathcal D)$.
BO selects new evaluations by maximizing an acquisition function $\alpha \colon \mathcal X\to\mathbb R$,
\begin{equation}
    x' \in \arg\max_{x\in\mathcal X} \alpha(x).
\end{equation}
Many improvement-based acquisitions are posterior expectations of a utility $u:\mathbb R\to\mathbb R$,
\begin{equation}
    \alpha(x)
    =
    \mathbb E_{p(Y(x)\mid\mathcal D)}
    \left[u(Y(x))\right].
\end{equation}
Expected improvement (EI) \citep{jones1998efficient} is the canonical example: for incumbent $f^\star=\max_{i\leq n}f(x_i)$, it uses the improvement utility $u(y)=(y-f^\star)_+$,
\begin{equation}
    \mathrm{EI}(x)
    =
    \mathbb E_{p(Y(x)\mid\mathcal D)}
    \left[(Y(x)-f^\star)_+\right].
\end{equation}
When evaluations can be performed in parallel, batch BO instead selects $q>1$ inputs that complement one another.

\subsection{Multi-objective Bayesian optimization}

In the multi-objective setting, the scalar objective is replaced by a deterministic vector-valued function $f=(f_1,\ldots,f_M):\mathcal X\to\mathbb R^M$, with all $M$ objectives maximized.
Accordingly, the surrogate represents uncertainty at $x$ through the random vector $Y(x)=(Y_1(x),\ldots,Y_M(x))\in\mathbb R^M$ with posterior predictive distribution $p(Y(x)\mid\mathcal D)$.
Objective vectors are only partially ordered: $y\in\mathbb R^M$ weakly dominates $y'\in\mathbb R^M$, written $y\succeq y'$, if $y_m\geq y'_m$ for every $m$, and dominates it, written $y\succ y'$, if at least one inequality is strict.
An input is Pareto optimal if its objective vector is not dominated by that of any other input, and the corresponding objective vectors form the Pareto front.
MOBO seeks a finite approximation of this front, which we denote by the nondominated set $R=\{r_1,\ldots,r_N\}\subset\mathbb R^M$.

A standard scalar measure of the quality of $R$ is its \emph{hypervolume}.
Given a reference point $z\in\mathbb R^M$, let
\begin{equation}
    D_z(R)
    :=
    \left\{
        y\in\mathbb R^M :
        z\preceq y
        \text{ and }
        y\preceq r
        \text{ for some }r\in R
    \right\}
    \label{eq:Dz}
\end{equation}
denote the region dominated by $R$ and bounded below by $z$, as illustrated in Figure~\ref{fig:rammobo-vs-hvi}.
Its hypervolume is
\begin{equation}
    \mathrm{HV}_z(R)
    :=
    \lambda\!\left(D_z(R)\right),
\end{equation}
where $\lambda$ denotes Lebesgue measure.
The hypervolume improvement (HVI) of a candidate objective vector $y$ is the additional dominated volume obtained by adding it to $R$,
\begin{equation}
    \mathrm{HVI}_z(y;R)
    :=
    \mathrm{HV}_z(R\cup\{y\})-\mathrm{HV}_z(R)
    =
    \lambda\!\left(D_z(R\cup\{y\})\setminus D_z(R)\right).
\end{equation}

HVI respects the Pareto order: if $y\succeq y'$, then $\mathrm{HVI}_z(y;R)\geq \mathrm{HVI}_z(y';R)$.
Moreover, it is \emph{strictly Pareto-monotone} on hypervolume-improving outcomes,
\begin{equation}
    y\succ y',
    \qquad
    \mathrm{HVI}_z(y';R)>0
    \quad\Longrightarrow\quad
    \mathrm{HVI}_z(y;R)>\mathrm{HVI}_z(y';R).
\end{equation}
This strictness is important when using a scalar utility to guide multi-objective search: if $y\succ y'$, assigning the same utility to both makes the acquisition insensitive to an objective improvement that comes at no cost in any other objective.
Some existing multi-objective utilities are only weakly Pareto-monotone and can therefore exhibit this behavior.
Our construction will instead preserve the strict Pareto monotonicity of HVI on hypervolume-improving outcomes.

Averaging HVI under the posterior predictive distribution gives expected hypervolume improvement (EHVI) \citep{emmerich2006single},
\begin{equation}
    \mathrm{EHVI}_z(x)
    :=
    \mathbb E_{p(Y(x)\mid\mathcal D)}
    \left[
        \mathrm{HVI}_z(Y(x);R)
    \right].
\end{equation}
This mirrors EI, with hypervolume improvement replacing scalar improvement as the realized utility.
With noisy observations $y_i=f(x_i)+\epsilon_i$, the latent objective vectors at previously evaluated inputs, and hence their dominance relationships, are uncertain.
Noisy expected hypervolume improvement (NEHVI) accounts for this uncertainty by averaging HVI jointly over the candidate outcome and the latent outcomes at the evaluated inputs \citep{daulton2021parallel}.
Writing $\mathcal R=\{Y(x_i)\}_{i=1}^n$, 
\begin{equation}
    \mathrm{NEHVI}_z(x)
    :=
    \mathbb E_{p(Y(x),\mathcal R\mid\mathcal D)}
    \left[
        \mathrm{HVI}_z\!\left(Y(x);\mathcal R\right)
    \right].
\end{equation}

\section{Random Max-Margin Multi-objective Bayesian Optimization}
\label{sec:methods}

RaMMOBO is built around a simple idea: replace expensive hypervolume improvement with the volume of a cheap geometric object that is guaranteed to lie inside the hypervolume-improvement region.
We obtain such an object from the maximin margin underlying expected maximin improvement (EMmI) \citep{svenson2016multiobjective}, whose evaluation scales linearly with the number of objectives $M$ for a fixed reference-set size.
The original margin, however, is reference-point-free and only weakly Pareto-monotone, so it does not retain key properties of hypervolume improvement.
We address these limitations in two steps.
First, we use the hypervolume reference point $z$ to turn the margin into an axis-aligned box whose volume lower-bounds HVI.
Second, we randomize the box geometry and average over these randomizations, recovering strict Pareto monotonicity on hypervolume-improving outcomes.
Each weighted utility evaluation costs $\mathcal O(NM)$ for $N$ reference outcomes, while the resulting acquisition provides a direct lower bound on (N)EHVI and supports efficient gradient-based acquisition optimization.

\subsection{Turning maximin improvement to an improvement volume}

We begin in the noiseless setting, with a fixed finite, nonempty nondominated set
$R=\{r_1,\ldots,r_N\}\subset\mathbb{R}^M$ and hypervolume reference point $z \in \mathbb{R}^M$.
For maximization, the EMmI acquisition of
\citet{svenson2016multiobjective} is
\begin{equation}
        \alpha_{\mathrm{EMmI}}(x)
        =\mathbb{E}_{p(Y(x)\mid\mathcal D)}\!\left[\Delta(Y(x);R)\right] \quad \text{with} \quad
        \Delta(y;R) = \min_{i=1,\ldots,N}\max_{m=1,\ldots,M}(y_m-r_{i,m})_+
    \label{eq:emmi}
\end{equation}
The margin $\Delta$ measures how far $y$ improves over every incumbent in
at least one objective. It is positive precisely when no point in $R$
weakly dominates $y$, but does not account for the hypervolume reference
point: a positive margin need not imply positive HVI.
In particular, $\Delta(y;R)$ can be positive even when $y_m\leq z_m$
for some objective $m$, in which case $y$ dominates no positive
$M$-dimensional volume above $z$ and therefore has zero HVI.

To connect this margin to hypervolume, we construct an axis-aligned
hyperrectangle (box) ending at $y$. Each side is bounded by $\Delta$ to exclude
the already-dominated region and clipped at $z$ to respect the hypervolume
reference point:
\begin{equation}
    \ell_m(y;R)=\min\!\left\{\Delta(y;R),(y_m-z_m)_+\right\},
    \qquad
    L(y;R)=\prod_{m=1}^M\ell_m(y;R).
    \label{eq:coordinatewise-box}
\end{equation}
For positive volume, clipping keeps this box above $z$, while the
maximin margin ensures that no incumbent weakly dominates its interior.
Its interior therefore lies in the hypervolume-improvement region, giving
\begin{equation}
    L(y;R)
    \leq\mathrm{HVI}_z(y;R).
    \label{eq:unweighted-box-bound}
\end{equation}
The bound is trivial at zero volume and is the unweighted special case
of the result proved in Appendix~\ref{app:hvi-bound}.

\subsection{Random weighting yields strict Pareto monotonicity}

The box utility $L$ can miss strict Pareto improvements because its max-margin construction may be controlled by a different coordinate.
For example, with $R=\{(0,0)\}$ and $z=(-3,-3)$, both $u=(2,1)$ and $v=(2,2)$ satisfy $L(u;R)=L(v;R)=4$ despite $v\succ u$.
Randomly changing the box aspect ratio removes this insensitivity by making different coordinates determine the box geometry under different weightings.

Specifically, for a positive weight vector $w\in\mathbb{R}_{>0}^M$, define the weighted margin
\begin{equation}
    \Delta_w(y;R)
    =
    \min_i \max_m w_m(y_m-r_{i,m})_+.
    \label{eq:weighted-margin}
\end{equation}
In the original objective coordinates, this margin corresponds to side length $\Delta_w/w_m$ in coordinate $m$.
Clipping these sides at the hypervolume reference point gives
\begin{equation}
    \ell_{w,m}(y;R)
    =
    \min\!\left\{
        \frac{\Delta_w(y;R)}{w_m},
        (y_m-z_m)_+
    \right\},
    \label{eq:weighted-box-lengths}
\end{equation}
and the resulting volume remains a pointwise lower bound on HVI,
\begin{equation}
    L_w(y;R)
    =
    \prod_{m=1}^M \ell_{w,m}(y;R)
    \leq
    \mathrm{HVI}_z(y;R).
    \label{eq:box-bound}
\end{equation}
Only the relative weights matter, since multiplying $w$ by a common positive constant leaves the box unchanged.
Setting $w=\mathbf{1}$ recovers the unweighted utility $L$, and Appendix~\ref{app:hvi-bound} proves the bound for general $w$.

We now average over random aspect ratios by drawing
\begin{equation}
    W\sim M\cdot\operatorname{Dirichlet}(\chg{\mathbf 1}),
    \qquad
    \bar L(y;R)
    =
    \mathbb{E}_W[L_W(y;R)].
    \label{eq:random-box-utility}
\end{equation}
This distribution treats objectives symmetrically and has support over all positive relative weightings.
While each fixed $L_w$ is only weakly Pareto-monotone, any strict Pareto improvement from an outcome with positive HVI is detected by a set of weightings with positive probability.
Averaging therefore recovers strict Pareto monotonicity while preserving the HVI lower bound:
\begin{equation}
    y\succ y',
    \qquad
    \mathrm{HVI}_z(y';R)>0
    \quad\Longrightarrow\quad
    \bar L(y;R)>\bar L(y';R).
    \label{eq:strict-pareto-monotonicity}
\end{equation}
The proof is given in Appendix~\ref{app:monotonicity}.

\begin{figure}[t]
  \centering
  \includegraphics[width=\textwidth]{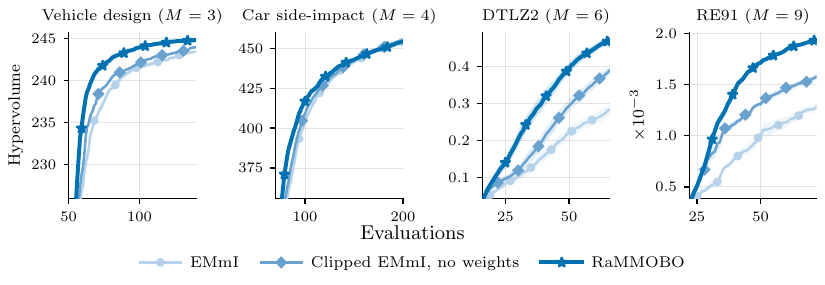}
  \vspace*{-0.5cm}
  \caption{\textbf{Impact of improvements to EMmI on common benchmark problems.} Starting from EMmI (Eq. \ref{eq:emmi}), we add clipping w.r.t. $z$, i.e., $\mathbb{E}_{p(Y(x)\mid\mathcal D)}[L(Y(x);R)]$ (Eq. \ref{eq:coordinatewise-box}), and finally introduce random weights to yield RaMMOBO. All three variants use the observed nondominated outcomes as a fixed reference set (noiseless setting) and share the same estimator and optimizer (Appendix~\ref{app:ablation-details}). RaMMOBO consistently improves hypervolume, where the gap becomes especially substantial when the number of objectives $M$ increases. Curves are the mean over 10 seeds with $\pm 1$ standard-error band.}
  \label{fig:emmi-improvements}
\end{figure}

\subsection{The RaMMOBO Acquisition Function}
 
\begin{algorithm}[t]
\caption{RaMMOBO's inner loop, sequential setting ($q=1$)}
\label{alg:rammobo}
\begin{algorithmic}[1]
\Require Data $\mathcal D=\{(x_i,y_i)\}_{i=1}^n$; surrogate with differentiable samples $Y(x)=g_{\mathcal D}(x,\xi)$; reference point $z$; number of Monte Carlo samples $S$; fatplus $\varphi_\tau$
\State $\hat{\mathcal X}\gets\{x_i:(x_i,y_i)\in\mathcal D,\ \mathbb E[Y(x_i)\mid\mathcal D]\text{ is nondominated}\}$
\State Draw and fix $w^{(s)}\sim M\cdot\mathrm{Dirichlet}(\mathbf 1)$ and independent base samples $\xi^{(s)},\xi^{(s)}_i$, \ $s\le S$, $x_i\in\hat{\mathcal X}$
\Function{$\hat\alpha$}{$x$} \Comment{estimates smoothed version of $\log\alpha_{\mathrm{RaMMOBO}}(x)$}
  \State $y^{(s)}\gets g_{\mathcal D}(x,\xi^{(s)})$,\ \ $r^{(s)}_i\gets g_{\mathcal D}(x_i,\xi^{(s)}_i)$ \Comment{for all $s$, $i$}
  \State $\Delta_w^{(s)}\gets\min_{i:x_i\in\hat{\mathcal X}}\max_m w^{(s)}_m\big(y^{(s)}_m-r^{(s)}_{i,m}\big)$
  \State $\ell^{(s)}\gets\sum_{m}\min\big\{\log\varphi_\tau(\Delta_w^{(s)})-\log w^{(s)}_m,\ \log(y^{(s)}_m-z_m)_+\big\}$ \Comment{$\log L_{w^{(s)}}$}
  \State \Return $\operatorname{logsumexp}_s\ell^{(s)}-\log S$
\EndFunction
\State \Return $x'\in\arg\max_{x\in\mathcal X}\hat\alpha(x)$ \Comment{multi-start L-BFGS-B, best quasi-random starts}
\end{algorithmic}
\end{algorithm}

In the noiseless setting, taking the posterior expectation of the randomized utility yields
\[
    \mathbb{E}_{p(Y(x)\mid\mathcal D)}
    \left[\bar L(Y(x);R)\right]
    \leq
    \mathrm{EHVI}_z(x).
\]
With noisy observations, the objective values at previously evaluated inputs are themselves uncertain, so the reference set must also be treated as random.
Following NEHVI, let
\[
    \mathcal R=\{Y(x_i)\}_{i=1}^n
\]
denote the latent objective vectors at the evaluated inputs.
Averaging $L_W$ jointly over the posterior of the candidate outcome $Y(x)$, the random reference set $\mathcal R$, and an independent weight vector $W$ defines the \emph{Random Max-Margin Multi-objective Bayesian Optimization} (RaMMOBO) acquisition:
\begin{equation}
    \boxed{
    \alpha_{\mathrm{RaMMOBO}}(x)
    =
    \mathbb{E}_{p(Y(x),\mathcal R\mid\mathcal D)}
    \mathbb{E}_W
    \left[
        L_W(Y(x);\mathcal R)
    \right]
    \leq
    \mathrm{NEHVI}_z(x)
    }.
    \label{eq:rammobo}
\end{equation}
The bound follows directly by averaging the pointwise inequality in Eq.~\chg{(\ref{eq:box-bound})} over the same joint posterior used by NEHVI \citep{daulton2021parallel}.

Crucially, the bound need not be tight to be useful: Since an acquisition function is used only through its maximizer, it suffices that the two functions correlate well: Empirically, we measure rank correlation and find strong agreement between RaMMOBO and $q$LogNEHVI for random inputs on the benchmark problems we consider; see Figure~\ref{fig:rank-agreement} and Appendix~\ref{app:rank-agreement}.

To form the reference set $\hat{\mathcal X}$ in practice, we use the subset of inputs $x_i$ in $\mathcal D$ whose posterior means $\mathbb E[Y(x_i)\mid\mathcal D]$ are nondominated.
As a second approximation, we sample $Y(x)$ and the selected reference outcomes independently from their posterior marginals for computational convenience.
The $q$LogNEHVI baseline instead prunes inputs using posterior samples and samples candidate and baseline outcomes jointly.

For numerical stability, we follow \citet{ament2023unexpected} and replace
the ReLU function in $\Delta_w(y;R)=[\min_i\max_m w_m(y_m-r_{i,m})]_+$ with the smooth
fatplus approximation. As shown in Algorithm~\ref{alg:rammobo}, we optimize the resulting Monte Carlo estimate
in log space using multi-start L-BFGS-B \citep{zhu1997lbfgsb}, reusing the weight draws $w^{(s)}$ and the random inputs to the posterior sampler throughout optimization. 
The NEHVI lower bound concerns the unsmoothed utility and exact expectation under the same joint posterior and full reference set as NEHVI; reference-set reduction, independent sampling, and smoothing are implementation approximations.

For batches ($q>1$), we select inputs one at a time.
\chg{The first input is selected using the single-candidate procedure above.
Each selected input then remains \emph{pending}: it belongs to the batch,
but its outcome has not yet been observed. To choose the next input, we
jointly sample possible outcomes at the pending inputs from the surrogate
posterior and add them to the reference set for each Monte Carlo sample.
Candidate outcomes are sampled conditionally on these pending outcomes,
preserving their posterior dependence. The acquisition therefore rewards
improvement beyond what the earlier batch members may already provide.
We repeat this process without collecting new observations or refitting
the surrogate.}
Once $q$ inputs are selected, we evaluate them in parallel
(Appendix~\ref{app:batch-rammobo}).

\section{Related Work}
\label{sec:related-work}
MOBO acquisition functions differ in how directly they target hypervolume improvement and in the computational machinery required to do so.
Scalarization-based methods such as ParEGO and its noisy batch variant $q$NParEGO apply improvement criteria to random augmented Chebyshev scalarizations \citep{knowles2006parego,daulton2020differentiable,daulton2021parallel}, avoiding explicit hypervolume computation but conditioning each acquisition value on sampled preference directions.
Relatedly, \citet{zhang2020random} express hypervolume as an expectation over random directions, which can be used as a partition-free Monte Carlo approximation to HVI.
Although inexpensive per sample, this approximation performs substantially worse than RaMMOBO on most benchmarks (Figure~\ref{fig:hv_diff_mc}), and its concentration guarantee has an exponential dependence on $M$; see Appendix~\ref{app:mc-hvi}.

Hypervolume-based methods retain the most direct improvement-based interpretation.
$q$NEHVI integrates hypervolume improvement over uncertainty in previously evaluated outcomes \citep{daulton2021parallel}, but exact hypervolume computation has super-polynomial worst-case scaling in $M$ \citep{yang2019efficient}.
$q$LogNEHVI improves the numerical behavior of this acquisition while retaining the same underlying geometric computation \citep{ament2023unexpected}.
Other approaches introduce different expensive subroutines: PESMO, MESMO, and multi-objective JES require auxiliary multi-objective optimization to construct posterior Pareto samples \citep{hernandezlobato2016predictive,belakaria2019maxvalue,tu2022joint}, $q$POTS applies NSGA-II to sampled objectives \citep{deb2002fast,renganathan2025qpots}, MOBO-OSD combines local Pareto-front estimation with hypervolume-based batch selection \citep{ngo2025moboosd}, and BOtied fits a multivariate CDF and optimizes its acquisition using CMA-ES \citep{park2024botied}.

Expected maximin improvement (EMmI) \citep{svenson2016multiobjective} is the construction most closely related to RaMMOBO because it also uses a partition-free margin that can be evaluated with cost linear in $M$.
However, its reference-point-free margin utility is only weakly Pareto-monotone and is not generally a lower bound on EHVI, while the original method treats observed outcomes as deterministic and relies on derivative-free acquisition optimization.
RaMMOBO retains the computational simplicity of this margin construction while turning it into a reference-point-aware volume utility, averaging over random objective weights, and integrating uncertainty in the reference outcomes.
This yields a partition-free lower bound on NEHVI that is strictly Pareto-monotone on hypervolume-improving outcomes and can be optimized with gradient-based methods, without box decompositions or inner multi-objective optimization.

\section{Experiments}
\label{sec:experiments}

\begin{figure}[t]
  \centering
  \includegraphics[width=\textwidth]{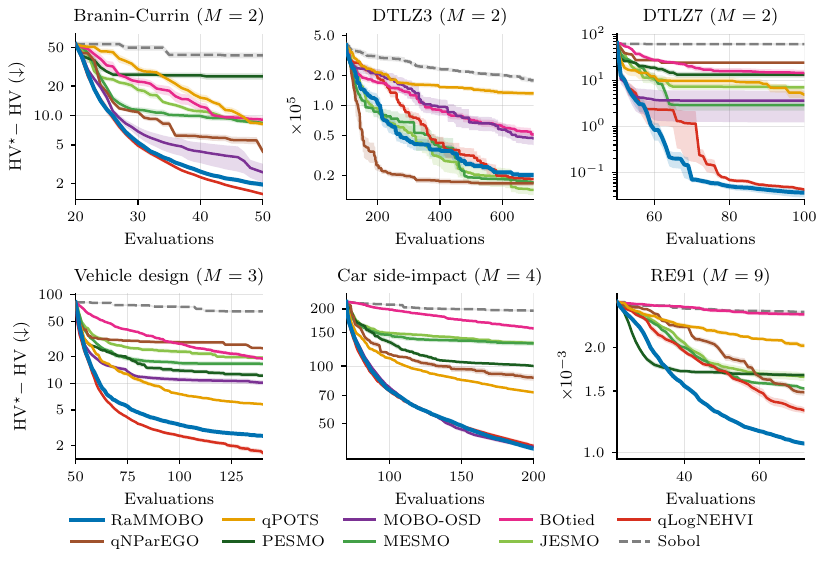}
  \vspace{-0.5cm}
  \caption{\textbf{Log hypervolume-difference against evaluations} on three synthetic and
  three real-world problems, $M = 2$ to $9$, in the sequential ($q=1$) setting.
  The $x$-axis counts function evaluations including the shared initial
  design. As exact box decompositions are computationally intractable at $M=9$, \chg{$q$}LogNEHVI uses approximate decompositions at $\alpha=10^{-4}$.}
  \label{fig:hv}
\end{figure}

In our experiments, we show that RaMMOBO becomes most advantageous for a large number of objectives, where hypervolume-based MOBO becomes computationally expensive.
Across synthetic and real-world benchmarks with $M=2$ to $14$, we observe competitive performance at lower objective counts, increasing gains over $q$LogNEHVI as $M$ grows, and acquisition optimization roughly two orders of magnitude faster than $q$LogNEHVI.
We evaluate sample efficiency and computational cost in noiseless and noisy, sequential ($q=1$) and batch ($q\in\{2,4\}$) settings.

\begin{figure}[tbp]
  \centering
  \includegraphics[width=\textwidth]{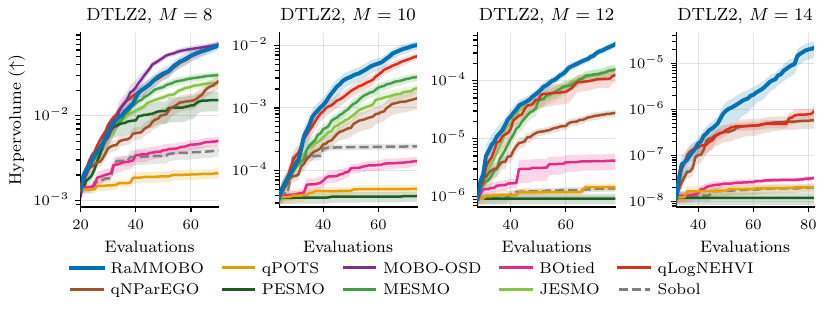}
  \vspace{-0.5cm}
  \caption{\textbf{Log hypervolume against evaluations for increasing number of objectives.} DTLZ2 at
  $M \in \{8, 10, 12, 14\}$, $2M + 4$ initial points and 50 rounds.
  As exact box decompositions are computationally intractable, \chg{$q$}LogNEHVI uses approximate decomposition at $\alpha = 10^{-3}$.}
  \label{fig:mstudy}
\end{figure}

\paragraph{Benchmark problems.}
We evaluate RaMMOBO on a mix of standard synthetic test functions and engineering design problems, spanning low- to many-objective regimes.
For the main comparison in Figure~\ref{fig:hv}, we use three synthetic and three real-world problems with $M=2$ to $9$ objectives.
The synthetic problems are Branin--Currin \citep{branin1972,currin1991} ($d=2$, $M=2$), DTLZ3 ($d=10$, $M=2$), and DTLZ7 ($d=5$, $M=2$) \citep{deb2005dtlz}.
The real-world problems are Vehicle \chg{design} \citep{liao2008vehicle} ($d=5$, $M=3$), Car side-impact \citep{jain2013carside} ($d=7$, $M=4$), and RE91 (RE9-7-1, car cab design; $d=7$, $M=9$) from the RE suite \citep{tanabe2020re}.
To isolate how performance changes with the number of objectives, we additionally evaluate DTLZ2 \citep{deb2005dtlz} at $M\in\{8,10,12,14\}$ with $d=M+1$ (Figure~\ref{fig:mstudy}).

\paragraph{Baselines.}
We compare against eight baselines spanning the main families of MOBO methods.
These include $q$LogNEHVI \citep{ament2023unexpected}, the numerically stabilized form of $q$NEHVI \citep{daulton2021parallel}; $q$NParEGO \citep{daulton2020differentiable}, the Monte Carlo batch variant of ParEGO \citep{knowles2006parego}; the information-theoretic methods PESMO \citep{hernandezlobato2016predictive}, MESMO \citep{belakaria2019maxvalue}, and JESMO \citep{tu2022joint}; the Thompson-sampling method $q$POTS \citep{renganathan2025qpots}; the search-direction-based MOBO-OSD \citep{ngo2025moboosd}; and the rank-based BOtied \citep{park2024botied}.
For MESMO and JESMO, we use the lower-bound implementations in BoTorch.
We additionally report scrambled Sobol sampling \citep{sobol1967,owen1998scrambling} as a no-model reference.

Reliable hypervolume baselines are particularly important for this comparison.
In our audit, the $q$NEHVI baseline reported by BOtied combines standardized outputs with an unstandardized reference point, while rerunning the released $q$POTS code does not reproduce the reported stagnation of its $q$NEHVI baseline on Branin--Currin.
We therefore use BoTorch's $q$LogNEHVI with consistent output and reference-point scaling throughout.

\paragraph{Experimental \chg{s}etup.}
All methods share the same surrogate, i.e., an independent GP per objective, with a Mat\'ern-$5/2$ kernel, inputs in the unit cube, standardized outcomes and inferred observation noise. Therefore, differences
between methods are differences between acquisition functions. Within a seed, every
method starts from the same initial design. In every experiment, we run 10 independent seeds and
report the mean with a $\pm 1$ standard-error band throughout. 
RaMMOBO does not use any problem-specific or $M$-dependent parameters. Experimental details can be found in Appendix~\ref{app:exp-details}.
Methods that exceed the available time or memory budget are omitted from
the corresponding panels; Appendix~\ref{app:baselines} details these limits.
For $M\geq8$, exact hypervolume decomposition is intractable, so $q$LogNEHVI
uses approximate decompositions: $\alpha=10^{-3}$ on DTLZ2 and
$\alpha=10^{-4}$ on RE91; Appendix~\ref{app:baselines} explains the
approximation parameter $\alpha$.

\begin{figure}[t]
  \centering
  \includegraphics[width=\textwidth]{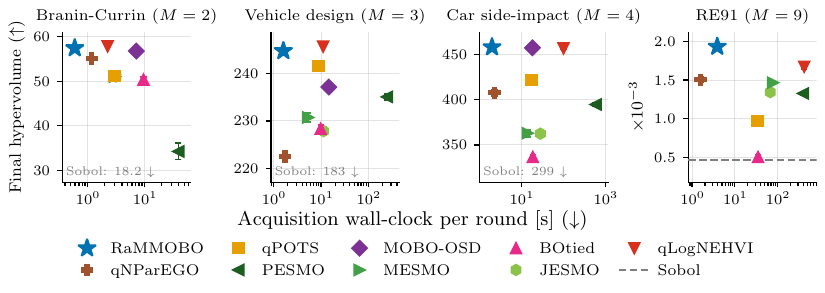}
  \vspace*{-0.5cm}
  \caption{\textbf{Hypervolume against wall-clock.} Final
  hypervolume against wall-clock time (seconds) needed to optimize an acquisition function to propose the next input (log $x$; lower-right is worse, upper-left is better). Sobol's acquisition is practically free\chg{; its final hypervolume is shown as a dashed line, or as a text label with a downward arrow where it lies below the plotted range}. On RE91, RaMMOBO's acquisition optimization is roughly two orders of magnitude faster than $q$LogNEHVI at $\alpha=10^{-4}$. See Appendix~\ref{app:additional-exps} for more panels.}
  \label{fig:cost}
\end{figure}

\paragraph{Metrics.}

We measure hypervolume \citep{zitzler1999hv} at a reference point fixed
before optimization and shared across methods. Figure~\ref{fig:hv} reports
$\mathrm{HV}^\star-\mathrm{HV}$ on a log scale (lower is better), where
$\mathrm{HV}^\star$ is an empirical reference value defined in
Appendix~\ref{app:metrics}; Figure~\ref{fig:mstudy} reports hypervolume itself
(higher is better). We also evaluate front coverage using the reference-point-free
IGD$^+$ metric \citep{ishibuchi2015igdplus} in Figure~\ref{fig:igd}.

\paragraph{Results and discussion.}

On the six benchmark problems (Figure~\ref{fig:hv}), $q$NParEGO improves
fastest on DTLZ3, while $q$LogNEHVI obtains the smallest final hypervolume
differences on Branin--Currin and Vehicle design. RaMMOBO performs comparably
to $q$LogNEHVI on Car side-impact and obtains the smallest final hypervolume
differences on DTLZ7 and RE91.
\chg{Figure~\ref{fig:emmi-improvements} isolates the effect of the two modifications to EMmI: adding clipping at $z$ and then random weighting improves hypervolume, most visibly at higher $M$.}
The IGD$^+$ results shown in Figure~\ref{fig:igd} also support RaMMOBO's strong Pareto front coverage, including an advantage over $q$LogNEHVI on Vehicle design.

As $M$ grows, Figure~\ref{fig:mstudy} shows that RaMMOBO is competitive at $M=8$ and achieves higher mean final hypervolume than approximate $q$LogNEHVI at $M\geq10$, with a widening gap at larger $M$: \chg{a}t $M=14$, it exceeds all shown competitors by more than an order of magnitude.

Figure~\ref{fig:cost} shows that this performance comes at substantially
lower computational cost. RaMMOBO achieves competitive final hypervolume
while optimizing its acquisition function orders of magnitude faster
than several competing methods (see Appendix~\ref{app:wallclock} for details).
These conclusions extend to Gaussian observation noise and to batch MOBO where $q\in\{2,4\}$
(Figures~\chg{\ref{fig:hv_diff_q2}, \ref{fig:hv_diff_q4}, and~\ref{fig:hv_diff_noisy}}).

\section{Conclusion}
\label{sec:conclusion}

We introduced RaMMOBO, which uses a simple acquisition function that makes
hypervolume-guided search computationally tractable as the number of objectives
grows. Its central idea is to measure max-margin box volumes under random
reweightings of the objective space. Averaging over weights and posterior
uncertainty yields a lower bound on (N)EHVI with evaluation cost linear in $M$, avoiding expensive box
decompositions. The utility is strictly Pareto-monotone on outcomes that improve
hypervolume, supports noisy observations, and admits efficient gradient-based
acquisition optimization. On synthetic and real-world benchmarks, RaMMOBO
matches or outperforms state-of-the-art methods, with the largest gains at
higher objective counts and acquisition optimization roughly two orders of
magnitude faster than $q$LogNEHVI on RE91. These findings demonstrate
that a tractable geometric lower bound can guide effective multi-objective
search even where directly computing expected hypervolume improvement becomes
prohibitive.

\section*{AI use statement}

The central idea of this work---lower-bounding hypervolume improvement by the
volume of a margin box, which is easy to compute---was developed by the authors without generative AI. We used
generative AI tools (large language models as coding and writing assistants)
for the following tasks with required disclosure: refining the theoretical
formulation, including the precise statements of the lower-bound and
monotonicity results, and assisting in writing their proofs
(Appendix~\ref{app:hvi-bound}--\ref{app:monotonicity}); implementing the
method and the experimental pipeline; translating the reference
implementations of qPOTS and MOBO-OSD into our code base; and giving feedback
on the experimental design and supporting the analysis of results. We have
not used generative AI tools to generate data sets, and qualitative or
thematic data analysis is not applicable to this work. Additionally, we used
generative AI tools for tasks with recommended disclosure: searching and
summarizing related literature, writing plotting code, and drafting and
editing parts of the paper.

We have reviewed all AI-assisted work. Each proof was checked line by line by
at least two authors, and the bounds were additionally verified numerically
against exact hypervolume computations on randomly generated instances.
AI-assisted code was reviewed by the authors, the transcribed
baselines were compared against the authors' reference implementations,
including reproductions of their published results where these were
available. All reported numbers were recomputed from the stored raw
experiment outputs, and all literature identified with AI assistance was
checked against the original sources. We take responsibility for the final
content of this work, including text, claims or artifacts produced with the
aid of generative AI.

\section*{Ethics statement}

This work is methodological: it proposes an acquisition function for
multi-objective Bayesian optimization and evaluates it on standard synthetic
test functions and publicly available engineering benchmark problems. It
involves no human subjects, personal or sensitive data, or newly released
data sets. As with machine learning and Bayesian optimization in general,
more efficient optimization methods can be applied to harmful ends, for
example to design materials, molecules, or systems with harmful properties;
our method does not specifically enable such uses, and responsibility for
the choice of objectives lies with those who apply it. We are not aware of
conflicts of interest, and we do not see ethical concerns specific to this
work that require further discussion.

\subsection*{Reproducibility statement}

RaMMOBO is specified in Section~\ref{sec:methods} and, as implemented,
in Algorithm~\ref{alg:rammobo} (sequential) and
Algorithm~\ref{alg:rammobo-batch} (batch). The assumptions and complete
proofs of the lower bound and of strict Pareto monotonicity are given in
Appendix~\ref{app:hvi-bound} and Appendix~\ref{app:monotonicity}.
Appendix~\ref{app:exp-details} describes every experiment: the benchmark
problems with their dimensions, budgets, reference points and noise levels,
the shared surrogate model, the settings of RaMMOBO and of every baseline,
the metrics, and the wall-clock measurement protocol. We release an
anonymized code repository at \url{https://anonymous.4open.science/r/rammobo}
that contains our implementation of RaMMOBO, all baselines as
run in our experiments, a version-pinned environment specification of all
software dependencies, and the configurations and scripts that reproduce every
experiment and figure in this paper.


\subsubsection*{Acknowledgments}
This research was funded in whole or in part by the Austrian Science Fund (FWF) (Standalone project ``TraCause'', 10.55776/PIN1217124) and the FWF Cluster of Excellence ``Bilateral AI'' (10.55776/COE12). 

\FloatBarrier
\bibliography{iclr2027_conference}
\bibliographystyle{iclr2027_conference}

\appendix
\section{Proofs}
\subsection{Hypervolume improvement lower bound}
\label{app:hvi-bound}
We prove the pointwise lower bound in Eq.~(\ref{eq:box-bound}).
Throughout, all objectives are maximized, $\preceq$ denotes componentwise
order, and $\lambda$ denotes the $M$-dimensional Lebesgue measure.

Let $R=\{r_1,\ldots,r_N\}\subset\mathbb{R}^M$ be a finite, nonempty
reference set, let $z\in\mathbb{R}^M$ be the hypervolume reference point,
and let $w\in\mathbb{R}_{>0}^M$. Recall the weighted margin and box lengths
from Eqs.~(\ref{eq:weighted-margin})--(\ref{eq:weighted-box-lengths}):
\begin{equation*}
    \Delta_w(y;R)=\min_i\max_m w_m(y_m-r_{i,m})_+,
    \qquad
    \ell_{w,m}(y;R)
    =\min\!\left\{\frac{\Delta_w(y;R)}{w_m},(y_m-z_m)_+\right\}.
\end{equation*}
Then, for every $y\in\mathbb{R}^M$,
\begin{equation}
    0\leq L_w(y;R)
    =\prod_{m=1}^M\ell_{w,m}(y;R)
    \leq\mathrm{HVI}_z(y;R).
    \label{eq:appendix-hvi-bound}
\end{equation}
The bound does not require $R$ to be nondominated or the weights to be
normalized.

\begin{proof}
Fix $y$. If $L_w(y;R)=0$, the claim follows from the nonnegativity of
hypervolume improvement.

Suppose $L_w(y;R)>0$. Then
$\varepsilon:=\Delta_w(y;R)>0$ and $y_m>z_m$ for every $m$, so
\begin{equation*}
    0<\ell_{w,m}(y;R)
    =\min\!\left\{\frac{\varepsilon}{w_m},y_m-z_m\right\}.
\end{equation*}
Consider the axis-aligned box
\begin{equation}
    B=\prod_{m=1}^M\bigl[y_m-\ell_{w,m}(y;R),y_m\bigr].
    \label{eq:appendix-certified-box}
\end{equation}
Since $\ell_{w,m}(y;R)\leq y_m-z_m$, every $u\in B$ satisfies
$z\preceq u\preceq y$. Hence $B\subseteq D_z(\{y\})$, with $D_z$
defined in Eq.~(\ref{eq:Dz}). For every $r_i\in R$, the definition of
$\varepsilon$ gives $\max_m w_m(y_m-r_{i,m})_+\geq\varepsilon$.
Choose a coordinate $m_i$ with
\begin{equation*}
    w_{m_i}(y_{m_i}-r_{i,m_i})_+\geq\varepsilon>0.
\end{equation*}
Strict positivity makes the positive part inactive in this coordinate, so
\begin{equation*}
    y_{m_i}-r_{i,m_i}
    \geq\frac{\varepsilon}{w_{m_i}}
    \geq\ell_{w,m_i}(y;R).
\end{equation*}
Every $u\in\operatorname{int}(B)$ therefore satisfies
$u_{m_i}>y_{m_i}-\ell_{w,m_i}(y;R)\geq r_{i,m_i}$, so $r_i$
cannot weakly dominate $u$. Since this holds for every reference outcome,
\begin{equation*}
    \operatorname{int}(B)
    \subseteq D_z(\{y\})\setminus D_z(R).
\end{equation*}
All sets above are measurable, since $D_z(R)$ is a finite union of
boxes. The boundary of $B$ has Lebesgue measure zero, so monotonicity of
measure gives
\begin{align*}
    L_w(y;R)
    &=\lambda(B)
      =\lambda\!\left(\operatorname{int}(B)\right)\\
    &\leq\lambda\!\left(D_z(\{y\})\setminus D_z(R)\right)
      =\mathrm{HVI}_z(y;R).\qedhere
\end{align*}
\end{proof}

Setting $w=\mathbf{1}$ recovers the unweighted bound in
Eq.~(\ref{eq:unweighted-box-bound}). Averaging over $W$ preserves the
inequality and yields $\bar L(y;R)\leq\mathrm{HVI}_z(y;R)$.
Taking the posterior expectation over the candidate outcome gives the
EHVI lower bound. In the noisy setting, the pointwise bound applies to
each realization of the candidate and reference outcomes. Averaging over
their joint posterior and an independent weight vector $W$ yields
\begin{equation*}
    \mathbb{E}_{p(Y(x),\mathcal R\mid\mathcal D)}\mathbb{E}_W\!\left[
        L_W(Y(x);\mathcal R)
    \right]
    \leq
    \mathbb{E}_{p(Y(x),\mathcal R\mid\mathcal D)}\!\left[
        \mathrm{HVI}_z(Y(x);\mathcal R)
    \right]
    =\mathrm{NEHVI}_z(x),
\end{equation*}
which establishes the acquisition bound in Eq.~(\ref{eq:rammobo}).

\subsection{Strict Pareto monotonicity}
\label{app:monotonicity}
We prove the strict Pareto monotonicity claim in
Eq.~(\ref{eq:strict-pareto-monotonicity}):
\begin{equation}
    y\succ y',\qquad \mathrm{HVI}_z(y';R)>0
    \quad\Longrightarrow\quad
    \bar L(y;R)>\bar L(y';R).
    \label{eq:appendix-strict-monotonicity}
\end{equation}

We work in the multi-objective setting $M\geq2$.
Here $R=\{r_1,\ldots,r_N\}\subset\mathbb{R}^M$ is a finite, nonempty
reference set, $z\in\mathbb{R}^M$ is the hypervolume reference point,
and $\bar L$ is the averaged box utility
\[
    \bar L(y;R)=\mathbb{E}_W[L_W(y;R)],
    \qquad
    W\sim M\cdot\operatorname{Dirichlet}(\chg{\mathbf 1}),
\]
with $\Delta_w$, $\ell_{w,m}$, and $L_w$ defined in
Eqs.~(\ref{eq:weighted-margin})--(\ref{eq:box-bound}).

\begin{proof}
Assume $y\succ y'$ and $\mathrm{HVI}_z(y';R)>0$.
Then $y'_m>z_m$ for every $m$: otherwise $D_z(\{y'\})$ has zero
$M$-dimensional measure. Also, no $r_i\in R$ weakly dominates $y'$, since
that would imply $D_z(\{y'\})\subseteq D_z(R)$.
Thus each reference has a positive gap at $y'$. For every $w>0$, its
weighted maximum is positive; finiteness of $R$ makes their minimum
$\Delta_w(y';R)$ positive as well. All box sides at $y'$ are therefore positive.

Only weight ratios matter. Indeed, for every $c>0$,
\[
    \Delta_{cw}(u;R)=c\,\Delta_w(u;R),
    \qquad
    \ell_{cw,m}(u;R)=\ell_{w,m}(u;R),
    \qquad
    L_{cw}(u;R)=L_w(u;R).
\]
We may therefore construct convenient unnormalized weights and normalize
them afterward without changing either box.

First, $L_w$ is weakly Pareto-monotone for every fixed $w>0$.
Indeed, $y\succeq y'$ implies
$(y_m-r_{i,m})_+\geq(y'_m-r_{i,m})_+$ for all $i,m$.
Multiplication by positive weights, followed by the maximum over $m$
and the minimum over $i$, preserves this inequality. Hence
$\Delta_w(y;R)\geq\Delta_w(y';R)$.
Since also $(y_m-z_m)_+\geq(y'_m-z_m)_+$, every box side is
nondecreasing:
\[
    \ell_{w,m}(y;R)\geq\ell_{w,m}(y';R).
\]
Hence
\begin{equation}
    L_w(y;R)\geq L_w(y';R)
    \qquad\text{for every }w>0.
    \label{eq:pointwise-weak-monotonicity}
\end{equation}

We next construct a weight vector for which the inequality is strict.
Choose a coordinate $k$ and set $\delta:=y_k-y'_k>0$.
It suffices to make the $k$th side strictly larger. Partition the reference
indices into
\[
    \chg{\mathcal I}=\{i:\exists m\neq k,\ y'_m>r_{i,m}\},
    \qquad
    \chg{\mathcal J}=\{1,\ldots,N\}\setminus \chg{\mathcal I}.
\]
The references in $\chg{\mathcal I}$ have a positive gap outside $k$; those in $\chg{\mathcal J}$ are
exceeded by $y'$ only in coordinate $k$.

We choose weights that elongate the unclipped box along the improved
coordinate $k$. Set $w_k=1$ and $w_m=C$ for $m\neq k$. There are only
finitely many positive gaps outside coordinate $k$, so we can choose
$C>1$ such that
\[
    C(y'_m-r_{i,m})_+>y_k-z_k
    \qquad\text{whenever }m\neq k\text{ and }y'_m>r_{i,m}.
\]
If no such gap exists, any $C>1$ suffices.

For $i\in \chg{\mathcal I}$, the weight choice and $y\succeq y'$ imply that its
weighted maximum at $y$ exceeds the clipping threshold $y_k-z_k$.
For $i\in \chg{\mathcal J}$, all gaps outside $k$ vanish at $y'$. Since $r_i$ does
not weakly dominate $y'$, we have
$y'_k>r_{i,k}$, and its weighted maximum at $y'$ equals $y'_k-r_{i,k}$.
This gap increases by $\delta$ at $y$, giving
\[
    \max_m w_m(y_m-r_{i,m})_+
    \geq y_k-r_{i,k}
    =y'_k-r_{i,k}+\delta
    \geq\Delta_w(y';R)+\delta.
\]
Taking the minimum over reference outcomes therefore gives
\[
    \Delta_w(y;R)
    \geq
    \min\!\left\{y_k-z_k,\Delta_w(y';R)+\delta\right\}.
\]
The clipping limit also increases by $\delta$, since
$y_k-z_k=y'_k-z_k+\delta$. Using $w_k=1$, clipping therefore yields
\[
    \ell_{w,k}(y;R)
    \geq
    \min\!\left\{y_k-z_k,\Delta_w(y';R)+\delta\right\}
    =\ell_{w,k}(y';R)+\delta.
\]
The $k$th side thus increases strictly. All other sides are nondecreasing
and all sides at $y'$ are positive, so
\[
    L_w(y;R)>L_w(y';R).
\]

Finally, normalize $w$ to $w^0=Mw/\sum_m w_m$ in the relative interior
of the normalized simplex,
\[
    \mathcal{S}_M^\circ
    =
    \left\{
        w\in\mathbb{R}_{>0}^M:
        \sum_m w_m=M
    \right\},
\]
preserving $L_{w^0}(y;R)>L_{w^0}(y';R)$ by scale invariance.

For fixed $y$ and $y'$, the difference $L_w(y;R)-L_w(y';R)$ is continuous in $w$ on
$\mathcal{S}_M^\circ$: its definition uses finite minima and maxima,
positive parts, products, and division by strictly positive coordinates.
The strict inequality therefore holds on a relatively open neighborhood
$U\subset\mathcal{S}_M^\circ$ of $w^0$. The distribution
$M\cdot\operatorname{Dirichlet}(\chg{\mathbf 1})$ has strictly positive
density throughout $\mathcal{S}_M^\circ$ with respect to the
$(M-1)$-dimensional measure on its affine hull, so
$\mathbb{P}(W\in U)>0$. Both expectations are finite, since
\[
    0\leq L_w(u;R)\leq\prod_{m=1}^M(u_m-z_m)_+
    \qquad\text{for every }w>0.
\]
The difference is nonnegative for every weight vector by
(\ref{eq:pointwise-weak-monotonicity}) and strictly positive on an event
of positive probability. Consequently,
\[
    \bar L(y;R)-\bar L(y';R)
    =
    \mathbb{E}_W\!\left[L_W(y;R)-L_W(y';R)\right]
    >0.\qedhere
\]
\end{proof}

\section{Batch-RaMMOBO}
\label{app:batch-rammobo}
Algorithm~\ref{alg:rammobo-batch} constructs a batch by selecting its
members greedily. Starting from an empty pending set $P$, each step
maximizes the acquisition against the evaluated reference inputs
$\hat{\mathcal X}$ and the inputs already in $P$, then adds the selected
input to $P$. The observed data, surrogate parameters, and reference
inputs remain fixed, while conditional predictions account for the
pending outcomes. The $q$ selected inputs are then evaluated in parallel.

For each Monte Carlo sample $s$, we draw the latent outcomes at the
pending inputs jointly and hold them fixed during the next acquisition
maximization. Write these outcomes as $R_P^{(s)}$. We draw the candidate
outcome conditional on $Y(P)=R_P^{(s)}$, giving the joint posterior law
$p(Y(x),Y(P)\mid\mathcal D)$ while keeping the pending outcomes fixed
as $x$ varies. \chg{In Algorithm~\ref{alg:rammobo-batch}, this is
implemented by applying the posterior sampler} $g_{\mathcal D}$
\chg{of Algorithm~\ref{alg:rammobo} jointly to $(x,P)$ with fixed base
randomness $\xi^{(s)}$.} Outcomes at the evaluated reference inputs
are sampled independently of one another and of the candidate and
pending outcomes. The sampled pending outcomes augment the reference
set, so each choice accounts for improvements the earlier choices may
already provide.

The weight draws and base randomness remain fixed during each inner
optimization. Pending outcomes are sampled for the updated set $P$
before the next greedy step; this requires no new observations or
surrogate refitting. When $P=\emptyset$, $R_P^{(s)}$ is empty and the
conditional candidate distribution reduces to $p(Y(x)\mid\mathcal D)$,
recovering the single-candidate procedure in Algorithm~\ref{alg:rammobo}.

\begin{algorithm}[t]
\caption{RaMMOBO's inner loop, batch setting ($q>1$)}
\label{alg:rammobo-batch}
\begin{algorithmic}[1]
\Require Data $\mathcal D$; surrogate with differentiable samples $Y(x)=g_{\mathcal D}(x,\xi)$; reference point $z$; batch size $q$; number of Monte Carlo samples $S$; fatplus $\varphi_\tau$
\State $\hat{\mathcal X}\gets\{x_i:\chg{(x_i,y_i)\in\mathcal D},\ \mathbb E[Y(x_i)\mid\mathcal D]\text{ is nondominated}\}$
\State Draw and fix $w^{(s)}\sim M\cdot\mathrm{Dirichlet}(\mathbf 1)$ and independent base samples $\xi^{(s)},\xi^{(s)}_i$, \ $s\le S$, $x_i\in\hat{\mathcal X}$
\Function{$\hat\alpha$}{$x;P$} \Comment{estimates smoothed $\log\alpha_{\mathrm{RaMMOBO}}(x)$ given pending points $P$}
  \State $\big(y^{(s)},\{r^{(s)}_p\}_{p\in P}\big)\gets g_{\mathcal D}\big((x,P),\xi^{(s)}\big)$ jointly;\ \ $r^{(s)}_i\gets g_{\mathcal D}(x_i,\xi^{(s)}_i)$, $x_i\in\hat{\mathcal X}$
  \State $\Delta_w^{(s)}\gets\min_{\chg{r\in\{r^{(s)}_i:\,x_i\in\hat{\mathcal X}\}\cup\{r^{(s)}_p:\,p\in P\}}}\max_m w^{(s)}_m\big(y^{(s)}_m-\chg{r_m}\big)$
  \State $\ell^{(s)}\gets\sum_{m}\min\big\{\log\varphi_\tau(\Delta_w^{(s)})-\log w^{(s)}_m,\ \log(y^{(s)}_m-z_m)_+\big\}$ \Comment{$\log L_{w^{(s)}}$}
  \State \Return $\operatorname{logsumexp}_s\ell^{(s)}-\log S$
\EndFunction
\State $P\gets\emptyset$
\For{$j=1,\ldots,q$} \Comment{greedy batch}
  \State $x'_j\in\arg\max_{x\in\mathcal X}\hat\alpha(x;P)$ \Comment{multi-start L-BFGS-B, best quasi-random starts}
  \State $P\gets P\cup\{x'_j\}$
\EndFor

\State \Return $P$
\end{algorithmic}
\end{algorithm}

\FloatBarrier

\section{Experimental Details}
\label{app:exp-details}

\subsection{Benchmark problems and protocol}
\label{app:problems}

All objectives are maximized; problems that BoTorch defines for minimization
are negated, together with their reference points. Table~\ref{tab:problems}
lists every problem. Branin--Currin, DTLZ3, DTLZ7, Vehicle \chg{design} and Car
side-impact are the sequential problems of \citet{renganathan2025qpots}, used
as implemented in BoTorch \citep{balandat2020botorch}, with an initial design
of $n_0=10d$ points and their numbers of iterations; note that
\citet{renganathan2025qpots} count iterations after the initial design, so
their evaluation budgets equal ours plus $n_0$. For DTLZ3 we report the first
600 of the 700 iterations. RE91 (car cab design) is our port of the RE
suite implementation \citep{tanabe2020re}, verified against it on 512
designs, with its four stochastic variables $x_8,\ldots,x_{11}$ held at their
means $(0.345, 0.192, 0, 0)$, so that the problem is deterministic and
observation noise is controlled by the protocol alone. The many-objective
study uses DTLZ2 \citep{deb2005dtlz} with $d=M+1$ (i.e.\ $k=2$ distance
variables) and follows \citet{park2024botied} in \emph{maximizing} its
objectives (BoTorch's DTLZ2 without negation). Its Pareto front is therefore
the part of the sphere of radius $1+\max g = 1.5$ in the positive orthant, not
the unit-sphere front of the minimization form. RE91 and DTLZ2 start from
$n_0=2M+4$ points and run 50 iterations.

\begin{table}[t]
\centering
\caption{Benchmark problems. $n_0$: initial design; $T$: iterations at $q=1$
(evaluations $n_0+T$); $z$: reference point, used both inside every
acquisition that takes one and for computing hypervolume (maximization
signs); $\sigma$: per-objective noise standard deviation in the noisy
setting.}
\label{tab:problems}
\setlength{\tabcolsep}{4pt}
\resizebox{\linewidth}{!}{%
\begin{tabular}{@{}lrrrrll@{}}
\toprule
Problem & $d$ & $M$ & $n_0$ & $T$ & reference point $z$ & noise $\sigma$ \\
\midrule
Branin--Currin & 2 & 2 & 20 & 30 & $(-18,\,-6)$ & $(3.02,\,0.126)$ \\
DTLZ3 & 10 & 2 & 100 & 600 & $(-10^4,\,-10^4)$ & $(16.0,\,16.7)$ \\
DTLZ7 & 5 & 2 & 50 & 50 & $(-15,\,-15)$ & $(0.010,\,0.174)$ \\
Vehicle \chg{design} & 5 & 3 & 50 & 90 & $(-1864.72,\,-11.82,\,-0.290)$ & $(0.377,\,0.051,\,0.002)$ \\
Car side-impact & 7 & 4 & 70 & 130 & $(-45.49,\,-4.51,\,-13.34,\,-10.39)$ & $(0.222,\,0.008,\,0.021,\,0.124)$ \\
RE91 & 7 & 9 & 22 & 50 & $-(41.30,\,1.03,\,0.90,\,0.78,\,1.47,$ & $(0.236,\,0.008,\,0.005,\,0.004,\,0.008,$ \\
 & & & & & $\;\;\;\,1.13,\,1.11,\,1.03,\,1.02)$ & $\;\;0.005,\,0.002,\,0.002,\,0.002)$ \\
DTLZ2 & $M{+}1$ & 6--14 & $2M{+}4$ & 50 & $-0.1\cdot\mathbf 1$ & --- (noiseless only) \\
\bottomrule
\end{tabular}}
\end{table}

\begin{figure}[t]
  \centering
  \includegraphics[width=\textwidth]{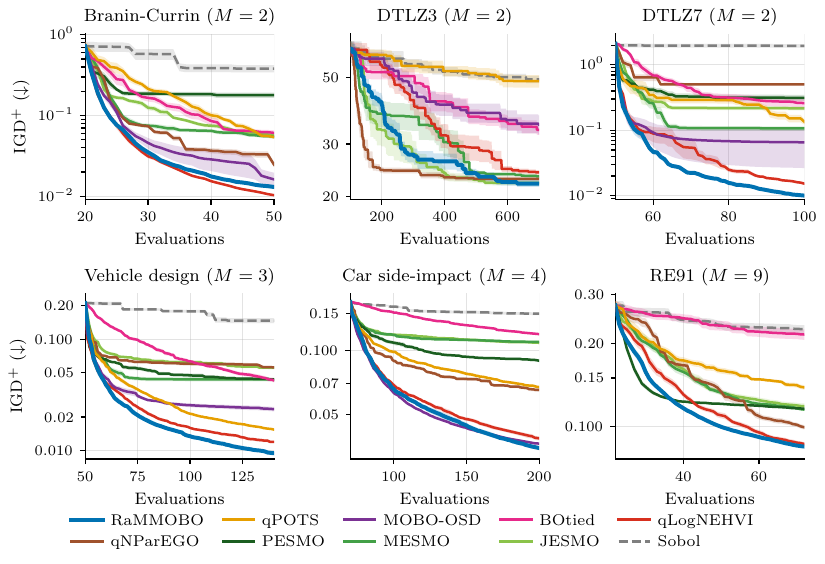}
  \vspace*{-0.5cm}
  \caption{\textbf{IGD$^+$ against evaluations} in the sequential ($q=1$) setting. Also in this reference-point-free metric, RaMMOBO performs on par with or outperforms state-of-the-art methods.}
  \label{fig:igd}
\end{figure}

\paragraph{Reference points.}
Hypervolume and every hypervolume-based acquisition depend on $z$, while
several baselines (qPOTS, qNParEGO, BOtied, PESMO, MESMO, JESMO, Sobol) do
not use one. To compare both groups fairly, we fix $z$ per problem before any
run, never tune it, and give every method that consumes a reference point
(RaMMOBO, qLogNEHVI, MOBO-OSD) exactly the $z$ that it is scored with.
Following \citet{daulton2021parallel}, $z$ is BoTorch's built-in point wherever
the problem defines one. For RE91 we take the per-objective worst value
over $2^{17}$ Sobol designs, rounded up (a design-space bound that uses no
knowledge of the Pareto front; its first coordinate lies slightly inside the
front's range of $f_1$). For DTLZ2 we use $0.1$ below the objectives' lower
bound of $0$. On Vehicle \chg{design} and Car side-impact, BoTorch derives $z$ from
an approximation of the true Pareto front \citep{tanabe2020re}; this informs
the methods that consume $z$ about the extent of the front and can favor them
over reference-point-free baselines on these two problems, which is why we also report a reference-free metric (IGD$^+$) in Figure~\ref{fig:igd}.

\paragraph{Noise.}
In the noisy setting, every observation, including the initial design, is
corrupted by independent Gaussian noise with the standard deviations of
Table~\ref{tab:problems}: $1\%$ of each objective's range, estimated as the
maximum minus the minimum over $2\cdot10^4$ Sobol designs. Hypervolume is
always computed from the noiseless objective values of the evaluated inputs.
The DTLZ2 study is noiseless.

\paragraph{Seeds and initial designs.}
Every configuration is run with 10 seeds. For each seed, the initial design
is a scrambled Sobol sample drawn after seeding all random number generators,
and is therefore identical across methods (including its noise). A curve is
drawn over the seeds that reach the problem's horizon; a method with fewer
than 5 such seeds is omitted from that panel.

\paragraph{Batch setting.}
For $q\in\{2,4\}$ we keep the evaluation budget $n_0+T$ of the sequential
setting and run $\lceil T/q\rceil$ iterations of $q$ points each; curves are
shown up to the largest whole batch within the budget (e.g.\ 48 of 50
evaluations on Branin--Currin at $q=4$). BoTorch-based acquisitions and
RaMMOBO (Algorithm~\ref{alg:rammobo-batch}) construct batches greedily with
the previously selected points as pending points; qNParEGO optimizes $q$
independently drawn scalarizations sequentially; qPOTS returns the $q$
points of its Pareto set approximation farthest from the observed inputs;
MOBO-OSD uses its kriging-believer batch selection; BOtied runs one CMA-ES
optimization per batch point.

\subsection{Surrogate model}
\label{app:surrogate}

All methods use the same surrogate: BoTorch's \texttt{SingleTaskGP} with an
independent GP per objective, a scaled Mat\'ern-$5/2$ kernel with one
lengthscale per input dimension, a constant mean, and a Gaussian likelihood
with inferred noise (also in the noiseless setting). Hyperparameters use
BoTorch's (v0.11.3) default priors: Gamma$(3,6)$ on lengthscales,
Gamma$(2,0.15)$ on the output scale, and Gamma$(1.1,0.05)$ on the noise
variance, which is bounded below by $10^{-4}$. Inputs are scaled to
$[0,1]^d$ and outcomes are standardized with the mean and standard deviation
of the current observations. The model is refitted from scratch in every
iteration by maximizing the marginal likelihood with L-BFGS-B (up to five
attempts on fitting errors). The reference point $z$ is expressed in the same
standardized units in every iteration. All computations use double
precision.

\subsection{RaMMOBO}
\label{app:rammobo-details}

We implement RaMMOBO as in Algorithms~\ref{alg:rammobo}
and~\ref{alg:rammobo-batch} with the same settings on every problem, for every
$M$, and in both noise settings: $S=128$ Monte Carlo samples, fatplus
temperature $\tau=10^{-3}$ (in standardized units), and one weight vector
$w^{(s)}\sim M\cdot\mathrm{Dirichlet}(\mathbf 1)$ per sample. The set
$\hat{\mathcal X}$ contains the observed inputs whose posterior means are
nondominated. Candidate and reference outcomes are sampled independently from
their posterior marginals using scrambled-Sobol normal base samples, which are
held fixed during acquisition optimization, as are the weights. The
acquisition is optimized with BoTorch's \texttt{optimize\_acqf}: 512
scrambled-Sobol candidates are evaluated, 10 starting points are selected by
BoTorch's Boltzmann heuristic (always including the best candidate), and each
is refined by L-BFGS-B \citep{zhu1997lbfgsb} for at most 200 iterations using
automatic differentiation.

\subsection{Baselines}
\label{app:baselines}

qLogNEHVI, qNParEGO, PESMO, MESMO and JESMO are BoTorch's implementations;
qPOTS and MOBO-OSD are transcribed from the authors' reference
implementations, while BOtied uses the authors' released implementation. All
run inside the same code base, on the surrogate of
Appendix~\ref{app:surrogate} and the same initial designs. Unless stated
otherwise, acquisitions are optimized with the same \texttt{optimize\_acqf}
settings as RaMMOBO (512 candidates, 10 restarts, L-BFGS-B with at most 200
iterations) and Monte Carlo acquisitions use 128 quasi-random samples.

\paragraph{qLogNEHVI} \citep{ament2023unexpected} uses baseline pruning and cached
Cholesky sampling. For $M<8$ it uses the exact box decomposition. For $M\ge 8$
the exact decomposition is intractable, and we use BoTorch's approximate
decomposition with relative threshold $\alpha$: $\alpha=10^{-4}$ on
RE91, and $\alpha=10^{-3}$ in the DTLZ2 study
(Figure~\ref{fig:mstudy}). Here, $\alpha$ is a relative cell-volume threshold:
cells overlapping the Pareto front are discarded once their volume is at
most an $\alpha$ fraction of the bounding-box volume enclosing the Pareto
set. Smaller $\alpha$ yields finer, more costly decompositions;
$\alpha=0$ recovers exact partitioning.
BoTorch's default, $\alpha=10^{M-8}$, yields no
cells at all for large $M$. At high $M$ the approximate decomposition of a
sampled front can also be empty in individual iterations, in which case
qLogNEHVI raises an error; we then assign zero improvement to all
candidates, as BoTorch's qNEHVI does, and the iteration selects a random
candidate. This happened in 0\%, 13\%, 46\% and 58\% of iterations at
$M=8,10,12,14$ in the DTLZ2 study. In the batch setting on RE91 we use
qLogNEHVI with $\alpha=10^{-4}$ and the same safeguard.

\paragraph{qNParEGO} \citep{daulton2020differentiable,knowles2006parego} draws Chebyshev
scalarization weights uniformly from the simplex (augmentation $0.05$), and
optimizes qNEI of the scalarized objective with baseline pruning.

\paragraph{PESMO, MESMO, JESMO}
\citep{hernandezlobato2016predictive,belakaria2019maxvalue,tu2022joint} use 10 sampled
Pareto sets of 10 points each, obtained by random search (population 2000, at
most 50 attempts) on posterior sample paths. MESMO and JESMO use BoTorch's
lower-bound variants with a box decomposition of each sampled front; PESMO
uses expectation propagation with BoTorch's defaults and is optimized without
gradients. If no valid set of Pareto samples is obtained within 100 attempts,
the iteration falls back to a Sobol point. On RE91 the lower bound of
MESMO and JESMO evaluates the log-determinant of a nearly singular matrix;
we compute it from eigenvalues floored at $10^{-6}$ to keep it finite. MESMO
and JESMO exceed the available memory (110\,GB) at $M=14$, and PESMO does
not reach the 600 iterations of DTLZ3 within our compute budget; they are
omitted there.

\paragraph{qPOTS} \citep{renganathan2025qpots} runs NSGA-II \citep{deb2002fast}
for 200 generations with population $100d$ on a posterior sample path and
selects the point(s) of the resulting Pareto set farthest from the observed
inputs. The authors' default of 10 generations did not reproduce their
reported results; 200 generations did.

\paragraph{MOBO-OSD} \citep{ngo2025moboosd} uses $n_\beta=20$ subproblems and
confidence multiplier $\delta=1.96$; each subproblem is solved by SLSQP from
four starts, and candidates are ranked by the hypervolume improvement of
their posterior mean at $z$. We fix four defects of the released code that
crash or stall the run and keep its selection of starting points as
released. Its per-iteration cost grows rapidly with $M$ (about 20\,h per run
at $M=8$; more than 8\,h for a single iteration at $M=10$), so it is run on
5 seeds at $M=8$ and omitted for $M\ge 9$.

\paragraph{BOtied} \citep{park2024botied} is the variant that ranks
candidates by a vine-copula CDF (nonparametric TLL family) of the posterior
means, optimized by CMA-ES (population 1024, 32 iterations, initial step size
0.2).

\paragraph{Sobol} evaluates one freshly scrambled Sobol point per iteration
\citep{sobol1967,owen1998scrambling}.

\subsection{Metrics}
\label{app:metrics}

\paragraph{Hypervolume difference.}
We compute the exact hypervolume \citep{zitzler1999hv} of the noiseless
objective values of all evaluated points at the problem's $z$
(Table~\ref{tab:problems}). $\mathrm{HV}^\star$ is the hypervolume of the
nondominated union of all points evaluated by any shown method, in both noise
settings and at all batch sizes, at the same $z$; the plots show
$\mathrm{HV}^\star-\mathrm{HV}$ on a log scale, with $\mathrm{HV}^\star$
shared by the sequential, noisy and batch figures. For RE91, computing
this union exactly is infeasible; we use a deterministic subset of 1200 of
its points that contains RaMMOBO's front, so its $\mathrm{HV}^\star$ is a
lower bound of the union's (yet above every method). The DTLZ2 study
(Figure~\ref{fig:mstudy}) shows hypervolume itself.

\paragraph{IGD$^+$} \citep{ishibuchi2015igdplus}.
For each problem, the reference front $Z$ is the nondominated union of the
noiseless values of all points evaluated by all methods in all runs (thinned
to 6000 points where larger), and every objective is normalized by the range
of $Z$. For an evaluated set $A$ we report
$\mathrm{IGD}^+(A)=\frac{1}{|Z|}\sum_{z'\in Z}\min_{a\in A}\|(z'-a)_+\|_2$
(maximization). IGD$^+$ uses no reference point.

\paragraph{Aggregation.}
Curves show the mean over seeds with a band of $\pm1$ standard error. The
$x$-axis counts evaluations including the initial design.

\subsection{Wall-clock time}
\label{app:wallclock}

Figure~\ref{fig:cost} plots the final hypervolume (mean $\pm1$ standard error
over 10 seeds, at the $z$ of Table~\ref{tab:problems}) against the time to
optimize the acquisition function once. We measure it by replaying each
method's recorded runs: for seeds 0 and 1 and six evenly spaced iterations
(e.g.\ $1,10,\ldots,50$ for 50-iteration problems) we restore the observed
data, refit the surrogate, and time the complete acquisition optimization,
including its setup (e.g.\ box decompositions or Pareto-set sampling), the
evaluation of the initial candidates and all optimizer restarts. Surrogate
fitting and function evaluations are excluded. All timings were taken on
identical compute nodes with two AMD EPYC 9354 CPUs (32 cores each,
$2\times 32$ cores / 128 hardware threads, 768\,GB RAM), on the CPU only (no
GPU), with each replay task limited to 4 threads and up to 15 tasks sharing a
node; we report the mean over iterations and seeds.

\subsection{Ablation}
\label{app:ablation-details}

The ablation (Figure~\ref{fig:emmi-improvements}) runs three acquisitions in
the noiseless setting on Vehicle \chg{design}, Car side-impact, DTLZ2 with $M=6$
($d=7$, $n_0=16$, 50 iterations, $z=-0.1\cdot\mathbf 1$) and RE91, with
the protocol of Table~\ref{tab:problems}. All three use the observed
nondominated outcomes as a fixed reference set $R$ (the noiseless setting of
Section~\ref{sec:methods}) and share RaMMOBO's estimator and optimizer:
fatplus with $\tau=10^{-3}$, averaging of $S=128$ samples in log space, and
the \texttt{optimize\_acqf} settings of Appendix~\ref{app:rammobo-details}.
\emph{EMmI} uses the margin $\Delta(y;R)$ of Eq.~(\ref{eq:emmi}) as its
utility; \emph{Clipped EMmI, no weights} uses the box volume $L(y;R)$ of
Eq.~(\ref{eq:coordinatewise-box}), i.e.\ $w=\mathbf 1$; and RaMMOBO uses
$L_W(y;R)$ with $W\sim M\cdot\mathrm{Dirichlet}(\mathbf 1)$.

\subsection{Rank agreement}
\label{app:rank-agreement}
\begin{figure}[t]
  \centering
  \includegraphics[width=\textwidth]{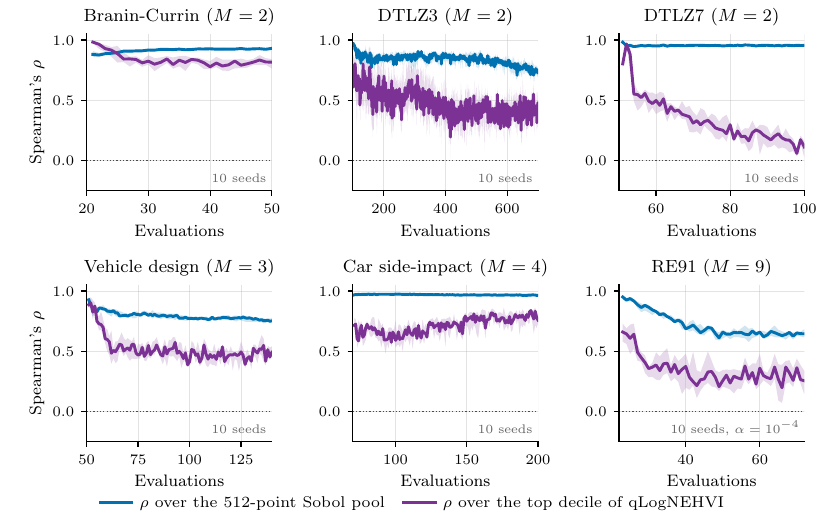}
  \vspace*{-0.5cm}
  \caption{\textbf{Rank agreement between RaMMOBO and $q$LogNEHVI} in the sequential ($q=1$) setting. Across the six benchmark problems, both acquisition functions rank a pool of 512 random inputs with consistently large Spearman correlation for $M \leq 4$ (blue curve), and their agreement weakens only on RE91 ($M=9$), where $q$LogNEHVI itself relies on an approximate box decomposition and is outperformed by RaMMOBO. The purple curve shows rank correlation on the top $10\%$ of points, ranked by $q$LogNEHVI; details in Appendix~\ref{app:rank-agreement}.}
  \label{fig:rank-agreement}
\end{figure}

To test whether RaMMOBO orders candidates as $q$LogNEHVI does, we rerun
RaMMOBO under the protocol of Table~\ref{tab:problems} (sequential,
noiseless, 10 seeds) with a \emph{shadow} $q$LogNEHVI attached: in every
iteration, after RaMMOBO's acquisition has been optimized and its candidate
chosen, $q$LogNEHVI is built on the same surrogate, the same observed data
and the same reference point $z$ (exact box decomposition for $M \le 4$;
$\alpha = 10^{-4}$ on RE91, as in Figure~\ref{fig:hv}), and both
acquisitions score the same pool of 512 scrambled-Sobol inputs drawn afresh
each iteration. We record Spearman's rank correlation between the two score
vectors over the whole pool and over the 51 inputs (top decile) that
$q$LogNEHVI ranks highest and show results in Figure~\ref{fig:rank-agreement}. Note that when the Pareto front has already been approximated well in later BO epochs, it may happen that the top decile of points all score similarly in terms of $q$logNEHVI, and the ranks may disagree with RaMMOBO only due to small changes in magnitude which are irrelevant for the BO performance: For example, RaMMOBO outperforms exact $q$logNEHVI on DTLZ7 in terms of hypervolume difference, but achieves bad top-decile rank agreement in later rounds (Figure~\ref{fig:rank-agreement}).
The shadow is never optimized and never influences
the run, so the trajectory is purely governed by RaMMOBO. Curves show the median
over the 10 seeds with the interquartile band.

\subsection{Software and hardware}
\label{app:software}

All experiments use BoTorch 0.11.3, GPyTorch 1.12 and PyTorch 2.5.1; qPOTS
uses pymoo 0.6.2 for NSGA-II, and BOtied uses pyvinecopulib 0.6.8 and cma
4.4.4. Runs were executed on CPU nodes with 2--4 threads per run.

\section{Additional Experiments}
\label{app:additional-exps}

\begin{figure}[t]
  \centering
  \includegraphics[width=\textwidth]{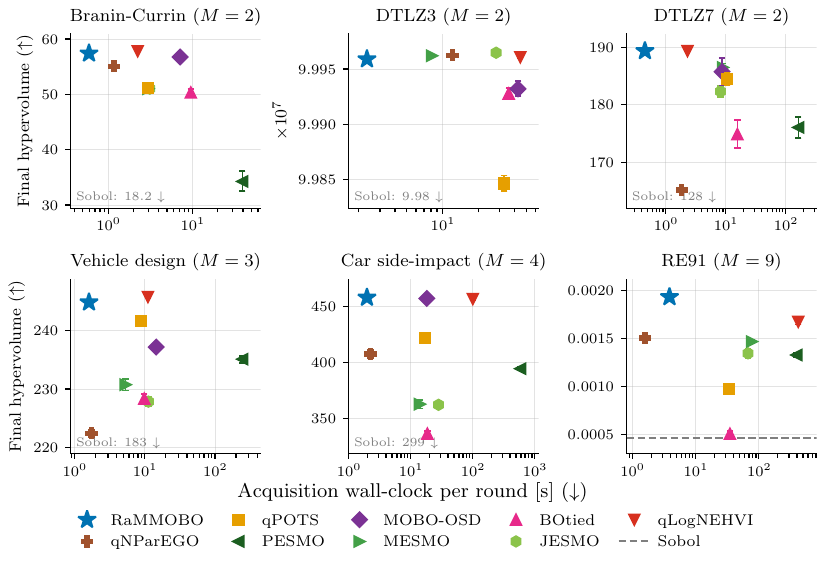}
  \vspace*{-0.5cm}
  \caption{\textbf{Hypervolume against wall-clock} ($q=1$). Final
  hypervolume against wall-clock time (seconds) needed to optimize an acquisition function to propose the next input. Same setup as in Figure~\ref{fig:cost}, but showing all six panels from the main set of experiments.}
  \label{fig:cost-2rows}
\end{figure}

\begin{figure}[t]
  \centering
  \includegraphics[width=\textwidth]{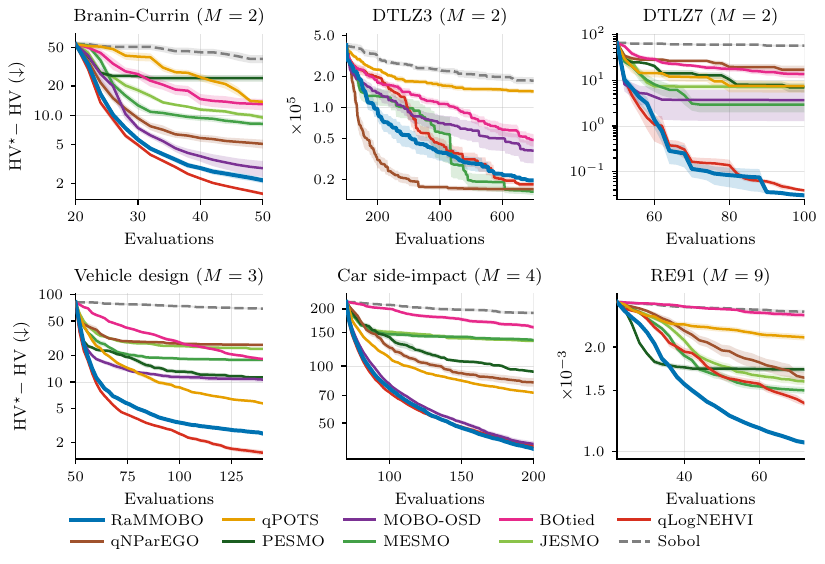}
  \vspace*{-0.5cm}
  \caption{\textbf{Log hypervolume difference} for $q=2$.
  We use batch RaMMOBO (Algorithm~\ref{alg:rammobo-batch}),
  which selects the two points greedily within each batch.
  Experimental details are given in Appendix~\ref{app:exp-details}.}
  \label{fig:hv_diff_q2}
\end{figure}

\begin{figure}[t]
  \centering
  \includegraphics[width=\textwidth]{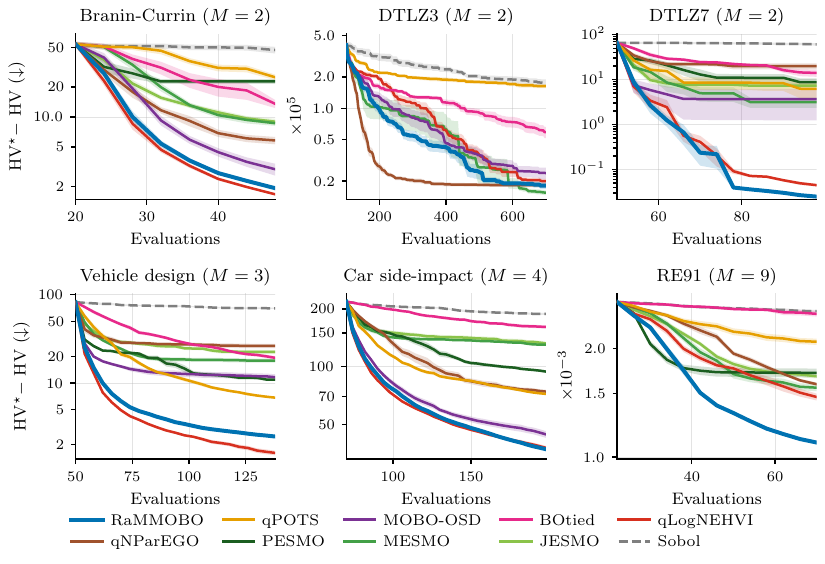}
  \vspace*{-0.5cm}
  \caption{\textbf{Log hypervolume difference} for $q=4$.
  We use batch RaMMOBO (Algorithm~\ref{alg:rammobo-batch}),
  which selects the four points greedily within each batch.
  Experimental details are given in Appendix~\ref{app:exp-details}.}
  \label{fig:hv_diff_q4}
\end{figure}

\paragraph{\textbf{Batch Setting.}}
Figures~\ref{fig:hv_diff_q2} and~\ref{fig:hv_diff_q4} repeat the six problems of Figure~\ref{fig:hv} with batches of $q=2$ and $q=4$ points, keeping the evaluation budget of the sequential setting (Appendix~\ref{app:problems}).
RaMMOBO is extended to $q>1$ with the same greedy scheme as the BoTorch-based baselines: each point of a batch is selected while treating sampled outcomes at the previously selected points as pending observations (Algorithm~\ref{alg:rammobo-batch}, Appendix~\ref{app:batch-rammobo}).

\begin{figure}[tbp]
  \centering
  \includegraphics[width=\textwidth]{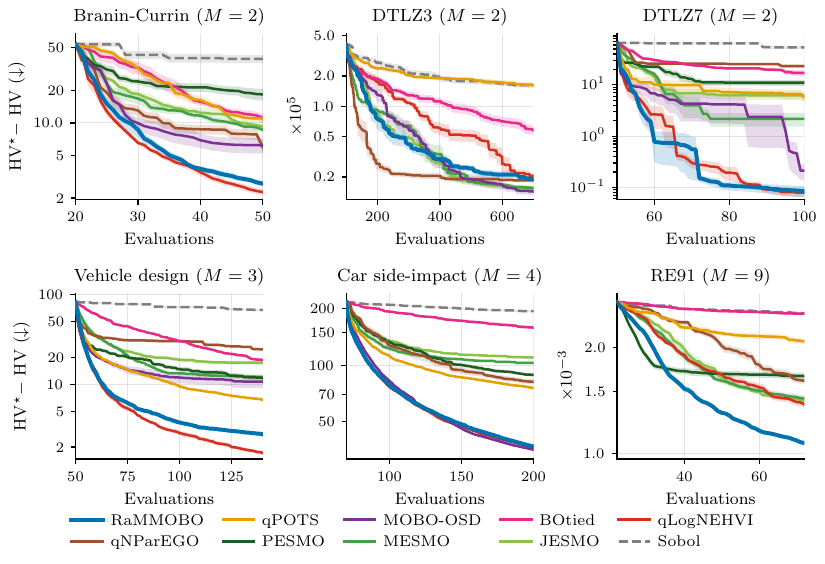}
  \vspace*{-0.5cm}
  \caption{\textbf{Log hypervolume difference} ($q=1$) in the noisy setting.
  Independent Gaussian noise with standard deviation equal to
  $1\%$ of each objective's range is added to every observation, as described
  in Appendix~\ref{app:exp-details}.
  Hypervolume is computed using the noiseless output of the black-box function.}
  \label{fig:hv_diff_noisy}
\end{figure}


\subsection{Monte Carlo hypervolume improvement}
\label{app:mc-hvi}
\begin{figure}[htbp]
  \centering
  \includegraphics[width=\textwidth]{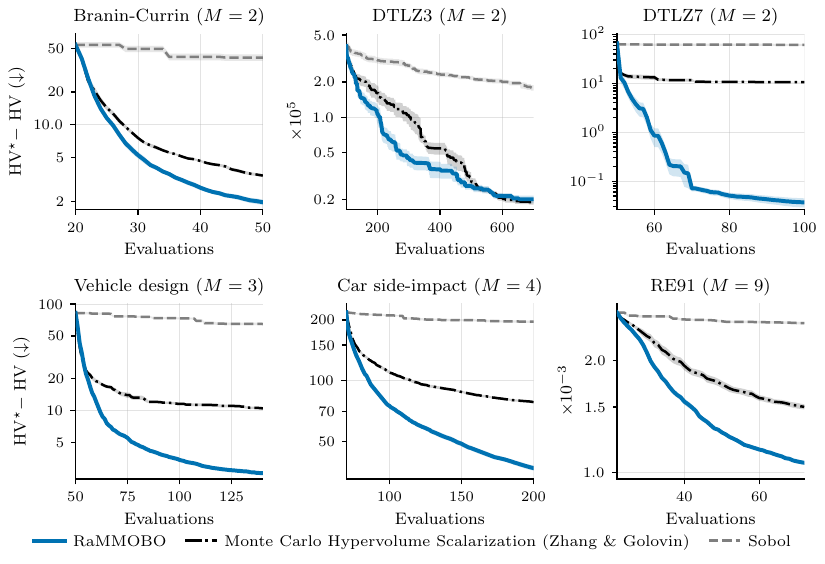}
  \vspace*{-0.5cm}
  \caption{\textbf{Log hypervolume difference} comparing RaMMOBO to a na\"ive Monte Carlo estimate of hypervolume's formulation as an expectation in \citet{zhang2020random}. We use identical experimental settings as in our  original $q=1$ run.}
  \label{fig:hv_diff_mc}
\end{figure}

The hypervolume identity of \citet[Lemma~5]{zhang2020random} expresses
hypervolume as an expectation over random scalarizations. Taking the
difference between the identities for $R\cup\{y\}$ and $R$ yields an
expectation for HVI, which can be estimated by averaging over sampled
directions without a box decomposition. Figure~\ref{fig:hv_diff_mc} compares
this na\"ive approximation with RaMMOBO under the same experimental protocol.
It produces larger hypervolume differences on most problems, with especially
poor progress on DTLZ7; on DTLZ3, it eventually reaches similar performance.
An unbiased estimate at a fixed candidate therefore need not yield an
effective acquisition optimizer at a finite sampling budget.

The concentration guarantee also retains an exponential dependence on the
number of objectives. With coordinate extents bounded by $B\geq1$, their
Lemma~6 gives the sufficient sample budget
\[
    K=\mathcal{O}\!\left(
        B^{2M}M^M\frac{\log(1/\delta)}{\varepsilon^2}
    \right)
\]
for additive error $\varepsilon$ in normalized hypervolume
$\mathrm{HV}/c_M$ with probability at least $1-\delta$, where
$c_M=\pi^{M/2}/(2^M\Gamma(M/2+1))$.
Thus this bound does not guarantee accurate approximation with a sample
budget polynomial in $M$. It is a sufficient bound, not a lower bound on
the necessary number of samples.

\end{document}